\documentclass[10pt,twocolumn,letterpaper]{article}

\usepackage[algorithms]{wacv}      

\usepackage{graphicx}
\usepackage{amsmath}
\usepackage{amssymb}
\usepackage{booktabs}
\usepackage[table]{xcolor} 
\usepackage[accsupp]{axessibility} 
\usepackage[pagebackref,breaklinks,colorlinks]{hyperref}

\usepackage[capitalize]{cleveref}
\crefname{section}{Sec.}{Secs.}
\Crefname{section}{Section}{Sections}
\Crefname{table}{Table}{Tables}
\crefname{table}{Tab.}{Tabs.}

\def\wacvPaperID{845} 
\def\confName{WACV}
\def\confYear{2024}
\def\datasetname{NITEC }

\begin{document}

\title{NITEC: Versatile Hand-Annotated Eye Contact Dataset for Ego-Vision Interaction}

\author{Thorsten Hempel, Magnus Jung, Ahmed A. Abdelrahman, and Ayoub Al-Hamadi \\
Neuro-Information Technology Group\\
Otto von Guericke University, Magdeburg, Germany\\
{\tt\small \{thorsten.hempel, magnus.jung, ahmed.abdelrahman, ayoub.al-hamadi\}@ovgu.de}
}

\maketitle

\begin{abstract}
Eye contact is a crucial non-verbal interaction modality and plays an important role in our everyday social life. While humans are very sensitive to eye contact, the capabilities of machines to capture a person's gaze are still mediocre. We tackle this challenge and present NITEC, a hand-annotated eye contact dataset for ego-vision interaction. 
NITEC exceeds existing datasets for ego-vision eye contact in size and variety of demographics, social contexts, and lighting conditions, making it a valuable resource for advancing ego-vision-based eye contact research. 
Our extensive evaluations on NITEC demonstrate strong cross-dataset performance, emphasizing its effectiveness and adaptability in various scenarios, that allows seamless utilization to the fields of computer vision, human-computer interaction, and social robotics. 
We make our NITEC dataset publicly available to foster reproducibility and further exploration in the field of ego-vision interaction\footnote{https://github.com/thohemp/nitec}.

\end{abstract}

\section{Introduction}
Eye contact plays a crucial role in our everyday social interactions and is one of the most important mechanisms in non-verbal interactions~\cite{specialeyecontact,farroni2002eye}.
It serves as signal to initiative for communication~\cite{ITIER2009843}, regulating interactions (e.g., establishing and maintaining joint attention~\cite{doi:10.1073/pnas.2106645118,Mauersberger2022ILA}) and to facilitate communication goals.
The effects of eye contact among humans are also observed in human-robot interaction scenarios.
In these settings, when individuals establish eye contact with a humanoid robot, it elicits similar types of automatic affective and attentional responses as they would during eye contact with another human ~\cite{KIILAVUORI2021107989, Zhang2017LookBD, doi:10.1126/scirobotics.abc5044}. 
Moreover, eye contact with a robot shows positive impact on its level of likability and attribution of human-likeness to a humanoid robot~\cite{kompatsiari} and can even effect a humans' honesty~\cite{10.3389/frai.2021.663190}. 

\begin{figure}
    \centering
    \includegraphics[width=\linewidth]{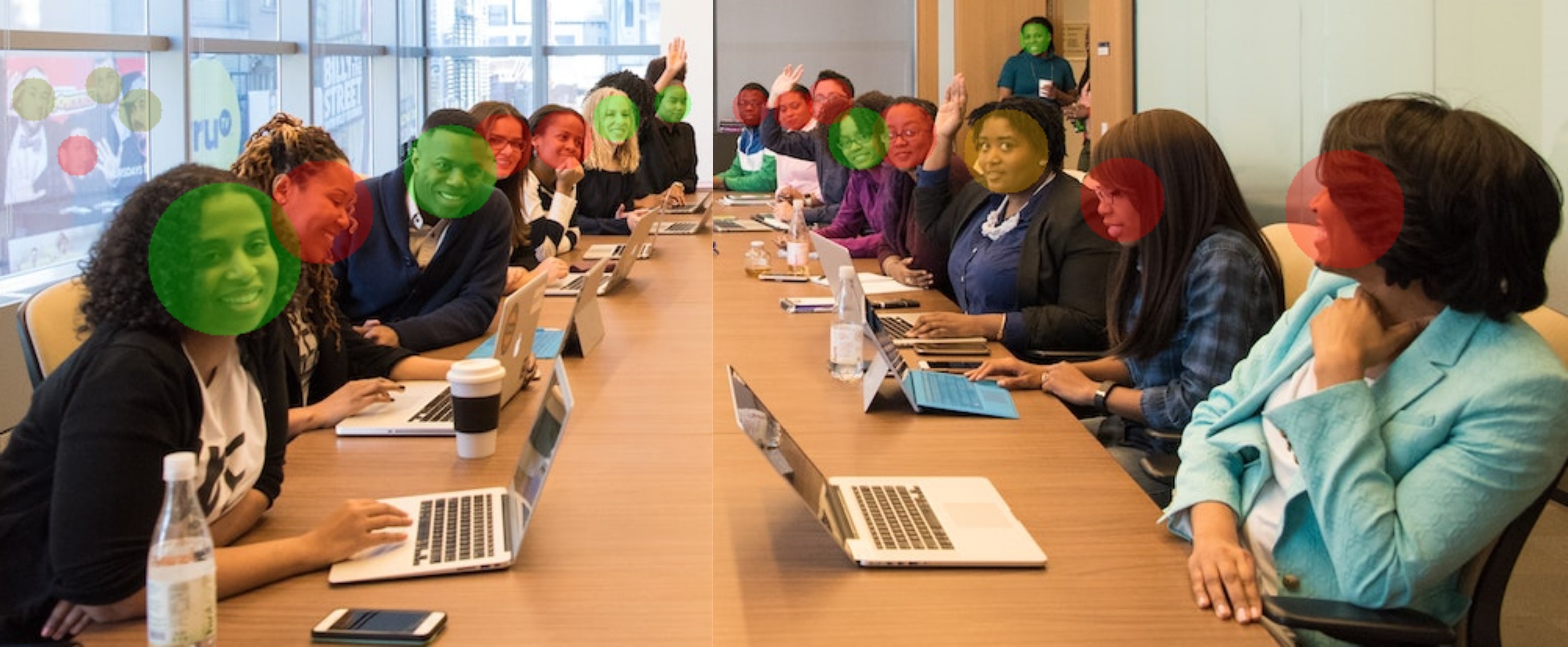}
    \caption{Typical discussion scenario in an office, where eye contact plays a crucial role to manage the interaction. With the aid of our NITEC dataset, machines are able to reach this communication level to achieve more intuitiveness in human-machine interactions. }
    \label{fig:enter-label}
\end{figure}

Humans possess a remarkable ability to perceive eye contact accurately, even in challenging conditions. However, the robust detection of eye contact in machines, particularly in the domain of human-robot interaction, has been largely unexplored, presenting a persistent and formidable challenge.
We argue that one of the main reasons for this is the lack of rich and high-quality datasets to effectively train neural networks to detect eye contact robustly in unconstrained settings. 

We strive to bridge this gap by introducing a new dataset, called NITEC, a dataset by the \textbf{N}euro-\textbf{I}nformation \textbf{T}echnology group for \textbf{E}ye \textbf{C}ontact detection in real life scenarios.
\datasetname provides hand-annotated labels for face-based eye contact detection from ego-perspective to target interaction scenarios in near- to midfield distances (\textit{e.g.}, see Figure~\ref{fig:enter-label}). It is based on four other datasets for different computer vision tasks, namely WIDER FACE~\cite{yang2016wider}, Gaze360~\cite{gaze360_2019}, CelebA~\cite{liu2015faceattributes}, and Helen~\cite{Le2012InteractiveFF}, that are partly re-annotated and combined to a new, well-curated dataset, that surpasses other current datasets in size, variety, and quality. In multiple experiments, we show that common CNN architectures trained on our dataset show striking generalization capabilities and outstrip other models on their own datasets. 
Further, we analyze the qualitative performance of our baseline models (with ResNet18 and ResNet50 backbone) and study the face area and the strictness for the classification of eye contact.
To summarize, our contributions are as follows:
\begin{itemize}
    \item We introduce and publicly release NITEC, a rich and large-scale eye-contact dataset for ego-vision interaction with 36,000 hand annotated samples.
    \item We evaluate our dataset in numerous quantitative experiments yielding state-of-the-art results using common classification models trained on NITEC
    \item We conduct qualitative evaluations to gain further insights about the spatial classification behavior and its corresponding consistency, highlighting the importance of eye contact prediction models  
\end{itemize}

\section{Related works}
There have been numerous research approaches tackling human-robot eye contact using dedicated gaze interaction systems~\cite{Sano2019EvaluatingIO, Kshirsagar2020RobotGB,Mishra2022KnowingWT,kompatsiari,9395532, Pan2020RealisticAI, Zaraki2014DesigningAE}. Most of these systems focus on realistic robot behavior, while the human's gaze is perceived by hardware-based Eye-Trackers~\cite{8956387,10.1145/2370216.2370368} that are not applicable to real life scenarios. Other image-based approaches focus mainly on gaze vector predictions or head pose estimation to identify the current focus of attention, where eye contact can be formulated as a subtask by defining the specific gaze angle~\cite{9794685}. However, we will show in the following sections that gaze predictions as well as head pose estimation models are barely sufficient to fulfill this task.

Eye contact as a classification task has recently drawn significant interest in the automotive area to estimate pedestrian's attention and awareness of the traffic situation.
Onkhar~\etal~\cite{10.1016/j.pmcj.2021.101455} presented a method for deriving eye contact in traffic using a head-mounted eye-tracker for the pedestrian and an in-vehicle stereo camera. Mordan~\etal~\cite{Mordan2020Detecting3P} introduced an end-to-end multi-task CNN for multi-attribute pedestrian analysis, including eye-contact, based on the JAAD dataset~\cite{8265243}. Another dataset called "LOOK" for pedestrian eye contact detection was introduced by Belkada~\etal~\cite{belkada2021pedestrians}, who proposed a body pose-based classification approach. This is caused by the fact that in the automotive application, most pedestrians are perceived from far distance, where faces alone cannot be captured with sufficient features.
Smith~\etal~\cite{10.1145/2501988.2501994} presented one of the early works for near- to midfield ego-vision eye contact classification based on a specifically created gaze datasets. Ye~\etal~\cite{Ye2015DetectingBF} presented another learning-based method, that couples a head pose-dependent appearance model with a temporal Conditional Random Field. But similar to gaze predictors, head-pose is not a reliable and precise eye contact indicator.
Chong~\etal~\cite{Chong2020DetectionOE} created a new image-based dataset with more than 4,000,000 samples to train neural networks that help identify the main gaze patterns for the diagnosis of Autism Spectrum Disorder. While their dataset remains non-public, Zhang~\etal~\cite{Zhang2021OnfocusDI} and Mitsuzumi~\etal~\cite{Mitsuzumi2017DEEPEC} published ego-vision based eye contact annotations for existing datasets along their model proposals. However, we will show that these datasets are not sufficiently sized and qualitative enough to build robust models upon them, leaving a gap for ego-vision based eye contact detection. We strive to close this gap by introducing NITEC, a manually annotated dataset 
, that encompasses various scenarios, diverse environments, and different difficulty levels.
\section{\datasetname\;Dataset}
In this section, we will give a detailed insight of the creation procedure and structure of our \datasetname dataset. We begin with a short analysis of existing datasets, followed by details of the collection of \datasetname and its annotation pipeline. Finally, we will give a short comparison of the final NITEC with other published datasets.

\subsection{Existing datasets}
To the best of our knowledge, there are only two publicly available datasets, that have been labeled with eye contact classes for near- and midfield detection range: DEEPEC~\cite{Mitsuzumi2017DEEPEC} and OFDIW~\cite{Zhang2021OnfocusDI}. DEEPEC provides manually annotated images provided by the datasets LFPW~\cite{fiducials_pami2013}, Helen~\cite{Le2012InteractiveFF}, AFW~\cite{koestinger2011annotated}, and IBUG~\cite{SAGONAS20163}, which sum up to 4,150 samples. The dataset is split into 53\% eye contact samples and 47\% samples with averted gaze. The source datasets are originally used for facial analysis and provide high-resolution, mostly non-occluded faces in unconstrained settings. OFDIW is split into an eye contact dataset for humans and eye contact dataset for animals. The human dataset consists of 16,548 samples with images collect from the LFW dataset~\cite{Huang2014LabeledFI}, which was originally published for face recognition tasks. A third --- not publicly available --- dataset was introduced by Chong~\etal~\cite{Chong2020DetectionOE}, who conducted a study with human subjects that resulted in 4,339,879 annotated frames (281,152 with eye contact) for training and 353,924 annotated frames (25,112 with eye contact) for validation. 

\subsection{Data composition}
Our objective was to create a comprehensive dataset for eye contact estimation in-the-wild, that provides large diversity and variability. Therefore, we selected publicly available images from four different datasets with complementary characteristics:
WIDER FACE~\cite{yang2016wider},
Gaze360~\cite{gaze360_2019},
CelebA~\cite{liu2015faceattributes} and
Helen~\cite{Le2012InteractiveFF}.
WIDER FACE is a large-scale in-the-wild dataset, primarily created for face detection tasks. It contains images with varying scene context including multiple persons, where the predominantly small resolution of faces makes the classification of eye contact particularly challenging. Gaze360 is a gaze estimation dataset capturing 238 subjects in indoor and outdoor environments. In sum, it provides 172,000 samples with a wide variety of gaze directions combined with a large range of head poses. CelebA is a large-scale celebrity face dataset that focuses on face attributes. This leads to images with feature-rich faces and challenging gaze directions (e.g., slightly next to the camera). Finally, Helen is another dataset for face feature analysis without celebrities setting providing various images gathered from flicker with extraordinary range of appearance variation, including pose, lighting, expression and occlusion. 

\subsection{Annotation procedure}
Our \datasetname dataset contains in total 35,919 hand-annotated samples, where 13,829 samples are from WIDER FACE, 7,214 are from Gaze360, 12,226 are from CelebA and 2,650 are from Helen. Except for Gaze360, all samples were manually annotated using a dedicated annotation tool that provides highlighted face crops based on a RetinaFace~\cite{deng2020retinaface} face detector. 
Thus, the annotators were able to incorporate the context outside the facial region into their decision-making process.
The annotators were asked to subjectively decide if the selected face appears to have eye contact with the camera/annotator or not. If uncertain, faces could be skipped and excluded from the dataset. 
The dataset has been split into 29,003 training images and 6,916 test images, leading to a split ratio of roughly 80/20. The labeling of the training set was distributed among two annotators, while for the test set, every sample was labeled by three annotators, where the majority vote determined the final label decision. Table~\ref{table:testset_annotation} gives an overview about the type of conflicts for each sub sets. It reports a conflict rate of roughly 15\% for WIDER FACE and CelebA, and about 8\% for Helen samples. Interestingly, in the latter two cases the majority vote determined in 90\% an eye contact sample, while for WIDER FACE most conflicts were selected to be no-eye contact. 

In contrast to the other datasets, Gaze360 provides 3D gaze vector annotations. We leverage this data, by converting the 3D gaze direction into a 2D vector, consisting of two angles \textit{yaw} and \textit{pitch}, and a unit vector for length. We then collected samples from the training and test set, where \textit{yaw} and \textit{pitch} would be between the strict thresholds of -5 and 5 degrees, indicating eye contact with the center of the camera. Likewise, we randomly sample the same number of samples with a gaze direction above the threshold for generating non-eye contact samples.

\begin{table}[t]
\begin{center}
\renewcommand{\arraystretch}{1.3}
\rowcolors{2}{gray!10}{gray!30}
\resizebox{\linewidth}{!}{%
\begin{tabular}{l cccc}
  \toprule
Dataset & No. of Samples & Conflicts [\%] & Eye contact in conflicts [\%] \\
\midrule
NITEC-WIDER FACE & 2829 & 15.6 & 24.7  \\
NITEC-CelebA & 2430 & 17.4 & 92.0 \\
NITEC-Helen & 525 & 8.2 & 88.4 \\
\midrule
NITEC & 5784 & 15.7 & 59.1  \\
\bottomrule
\end{tabular}}
\caption{Evaluation of the mutual annotated test sets by three annotators with the number of annotated samples, the share of annotation conflicts and share of label decision based on majority vote.}
\label{table:testset_annotation}
\end{center}
\end{table}

\subsection{Dataset comparison}
\begin{figure}[t]
    \centering
\begin{subfigure}{0.49\linewidth}
    \includegraphics[width=\linewidth]{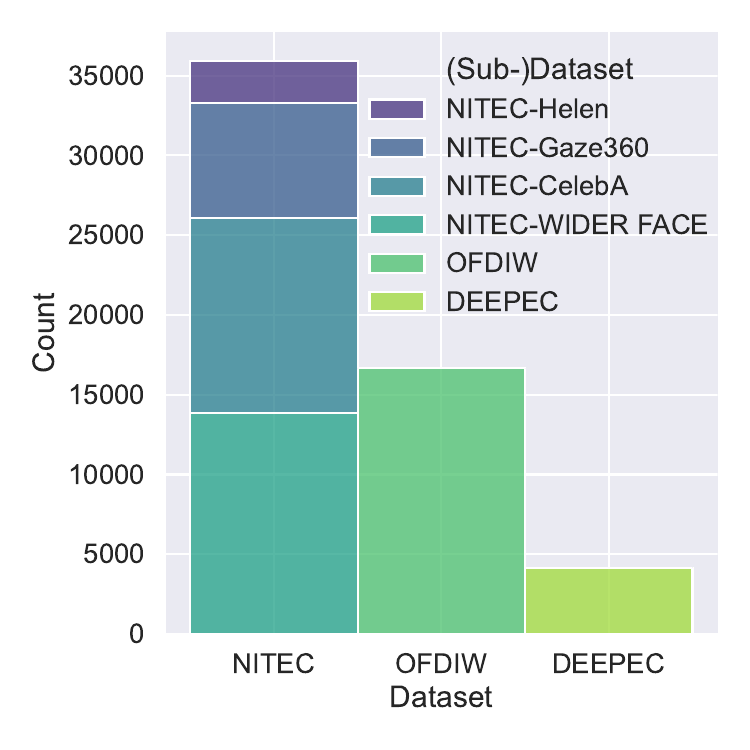} 
\end{subfigure}
\begin{subfigure}{.49\linewidth}
    \includegraphics[width=\linewidth]{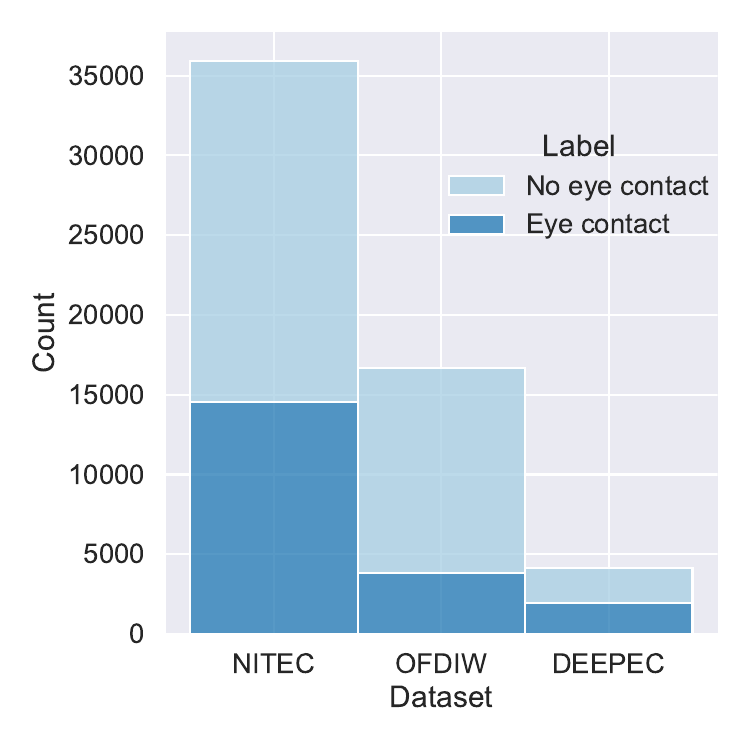}
\end{subfigure}
\caption{Comparison of our proposed dataset with two other public datasets in size and label distribution.}
\label{fig:comparison_of_public_datasets}
\end{figure}

Our \datasetname dataset will be freely available and will include the precise position of the facial region in the original image, as well as the annotations provided by the annotators. 
The test scripts, such as for Figure 5, will also be provided for use in research. 
Table~\ref{table:dataset_comparison} and Figure~\ref{fig:comparison_of_public_datasets} show a comparison of our proposed dataset with the two other public datasets. With around 36,000 samples, our NITEC dataset is more than double the size of OFDIW that contains 16,648 samples. The third dataset, DEEPEC, consists only of 4,150 samples and is therefore the smallest one. However, with a label split of 47.5\% eye contact samples and 52.5\% it is the most balanced candidate, followed by our NITEC dataset with 40.5\% eye contact. The slight overhang of non-eye contact samples is introduced by the WIDER FACE subset, where only approximately every fifth sample is labeled with eye contact, based on the subset ratio. This is caused by the nature of the WIDER FACE dataset, that contains mainly faces captured from far distances, where the person is not aware of the camera. We chose this dataset with the intention to reduce false positive in the target models for cases where target faces are feature-poor. The remaining NITEC subsets are fairly balanced around 50\%. However, the OFDIW dataset has a similar label distribution compared to our WIDER FACE subset with unbalanced 23\%.

\begin{table}[t]
\begin{center}
\renewcommand{\arraystretch}{1.3}
\rowcolors{2}{gray!10}{gray!30}
\resizebox{\linewidth}{!}{%
\begin{tabular}{l ccc}
  \toprule
Dataset&\multicolumn{3}{c}{No. of Samples (Eye Contact [\%])}\\
  \cmidrule{2-4}
 &Train &  Test& $\sum$\\ 
\midrule
OFDIW & 11,511 [22.7] & 4,137 [23.93] &\multicolumn{1}{c}{16,648 [23.0]}  \\
DEEPEC & \multicolumn{2}{c}{4,150 [47.5]}& \multicolumn{1}{c}{4,150 [47.5]} \\ \midrule
NITEC-WIDER FACE(Ours) & 11,000 [23.2] & 2,829 [20.0] & 13,829 [22.6] \\
NITEC-CelebA (Ours) &  9,829 [49.5] & 2,397 [63.4] &  12,226 [52.2]\\
NITEC-Helen (Ours) &  2,125 [52.0]  & 525 [57.1] & 2,650 [53.0]\\
NITEC-Gaze360 (Ours) & 6,049 [50.5] & 1,165 [50.6] & 7,214 [50.5] \\
\midrule
NITEC (Ours) &   29,003 [39.9] & 6,916 [43.0] & \textbf{35,919} [40.5] \\
\bottomrule
 \end{tabular}}
 \caption{Comparison of our proposed dataset with other public eye contact datasets with corresponding label distribution.}
 \label{table:dataset_comparison}
\end{center}
\end{table}

\begin{table*}[ht]
\begin{center}
\renewcommand{\arraystretch}{1.3}
\rowcolors{2}{gray!10}{gray!30}
\resizebox{\textwidth}{!}{%
\begin{tabular}{ll ccccccc}
  \toprule
 Training Dataset  & Method&  \multicolumn{7}{c}{Eye Contact Classification (AP) ↑ [F1-Score ↑] } \\\cmidrule{3-9}
 & & OFDIW & DEEPEC & NITEC-WF~\cite{yang2016wider} & NITEC-Gaze360~\cite{gaze360_2019} & NITEC-CelebA~\cite{liu2015faceattributes} & NITEC-Helen~\cite{Le2012InteractiveFF} & NITEC\\
\cmidrule{1-9}
OFDIW~\cite{Zhang2021OnfocusDI} &   ResNet18 & 57.4 [33.1] & 70.8 [61.2] & 44.3 [37.2] & 76.3 [19.9] &  91.6 [76.7] & 92.1 [75.4] &    80.4 [61.2]\\
DEEPEC~\cite{Mitsuzumi2017DEEPEC} & ResNet18 & 31.2 [16.3] & 69.6 [62.7] &27.4 [23.9] & 57.8 [27.7] &   80.4 [42.1] &  88.6 [74.5]  &      62.0 [39.9]\\
NITEC (Ours) &                      ResNet18 &   \textbf{59.5 [55.3]} & \textbf{72.4 [73.3]} & \textbf{57.0 [59.8]}  & \textbf{93.0 [86.6]} & \textbf{96.0 [90.2]} & \textbf{95.6 [89.5]}  &  \textbf{88.9 [83.6]} \\ \cmidrule{1-9}
OFDIW~\cite{Zhang2021OnfocusDI} &   ResNet50 & 55.2 [40.0] & 68.8 [59.3] & 41.3 [38.7] &  68.5 [19.2] &  90.4 [73.7] & 90.1 [72.7] &    75.8 [59.0]\\
DEEPEC~\cite{Mitsuzumi2017DEEPEC} & ResNet50 & 31.2 [10.7] & \textbf{75.7} [65.8] & 26.3 [17.7] &  54.3 [22.5] &   83.1 [38.8] &  93.0 [74.5]  &      63.1 [37.1]\\
NITEC (Ours) &                       ResNet50 & \textbf{57.2 [53.6]} & 73.8 \textbf{[71.7]} & \textbf{57.2 [57.0]} & \textbf{89.7 [84.5]} & \textbf{95.2 [90.6]} & \textbf{96.7 [90.5]} & \textbf{87.9 [83.0]}\\

\bottomrule
\end{tabular}}
\caption{Comparison of different datasets using simple baseline models based on ResNet18 and ResNet50. Eye contact classification is evaluated using the average precision (AP) metric and F1-Score. Each model is trained as a simple classifier for 20 epochs on the training set of each dataset and tested on all other combinations of test sets.}
\label{table:ComparisonOfSimpleBaselineModels}
\end{center}
\end{table*}

\section{Experiments}
We conduct several experiments to analyze the performance and quality of our NITEC dataset. We begin with a quantitative analysis by comparing the performance of baseline models over multiple dataset to study the cross-dataset generalization. In a second experiment, we compare these models with other eye contact detection models and other gaze prediction and head pose based approaches. Finally, we conduct an intra-dataset experiment to evaluate the impact of each of our NITEC datasets component on the overall performance. While for our NITEC and OFDIW the train and test sets are predefined, there is no definition for DEEPEC. Therefore, we randomly split DEEPEC into 80/20 ratio for training and testing. In additional qualitative evaluations, we analyze the performance on different exemplary images and assess the prediction pattern.

\subsection{Experimental setup}
To train our baseline models for the binary classification between eye contact and non-eye contact, we chose the popular and simple ResNet~\cite{He_2016_CVPR} and SWIN-Transformer~\cite{liu2021Swin} backbone with two output neurons. 
The input consists of the cropped faces, and we limit the augmentation to random cropping and random horizontal flip.
The model is trained for 20 epochs, using binary cross-entropy loss function with Adam optimizer~\cite{kingma2014adam}, with a learning rate of 0.0001 (0.001 for the SWIN-Transformer) and a batch size of 80. 
This way, we obtain a standardized model with focus on the training data to enable optimal comparisons.

\subsection{Cross-dataset evaluation}
\label{sec:cross_eval}
Comparing the available datasets specifically designed for eye contact detection (OFDIW, DEEPEC, and our NITEC dataset), the models were trained on the respective training sets of each dataset, and their results were compared on all test sets of the datasets and sub-datasets shown in table~\ref{table:ComparisonOfSimpleBaselineModels}.
We follow the strategy by Belkada~\etal~\cite{belkada2021pedestrians} and employ the average precision as the main metric, complemented by the F1-Score, to provide insights into the classification model's ability to accurately classify positive instances while minimizing false positives and false negatives.
Examining the results of ResNet18 reveals a clear distinction between models trained on different training datasets. 
DEEPEC consistently performs the worst on all test datasets, showing significant differences compared to the other models (which could be attributed to the size of the training dataset). 
The OFDIW ResNet18 model also performs consistently worse than the NITEC ResNet18 model, even on the test data of the OFDIW dataset itself.
Given the relatively poor performance of all models on the OFDIW dataset, it is likely that the OFDIW dataset is plagued by significant label noise.
Analyzing the test datasets where the differences between models are most prominent, it is evident that the results on the particularly challenging WIDER FACE dataset not only perform worse compared to other test datasets but also exhibit significant variations between models, with the NITEC model  consistently outperforming the others. 
Similarly, on the challenging CelebA dataset, which contains difficult gaze angles passing closely by the camera. 
As these results hold for both average precision and F1-score, the NITEC dataset enables better generalization of the relevant features for eye contact data compared to the other datasets.
These findings can be extended to the more complex ResNet50 models. 
Here, too, the NITEC-trained model outperforms the others, except for the DEEPEC test dataset, where the model trained on DEEPEC data achieves a slightly better average precision, while the F1-score remains higher for the NITEC model.
Comparing the ResNet18 models with the ResNet50 models reveals that the larger models do not yield significant improvements (except for a slight improvement in DEEPEC), but rather lose robustness. 
It is assumed that the ResNet50 models overfit on the datasets of this size due to faster convergence within 20 epochs.

\begin{figure}
\rotatebox[origin=l]{90}{ResNet18}
\begin{subfigure}{.30\linewidth}
    \includegraphics[width=\linewidth, trim=0 15 0 10, clip]{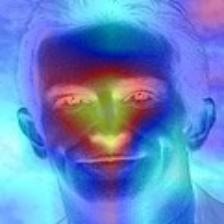}
\end{subfigure}
\begin{subfigure}{.30\linewidth}
    \includegraphics[width=\linewidth, trim=0 15 0 10, clip]{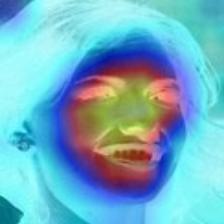}
\end{subfigure}
\begin{subfigure}{.30\linewidth}
    \includegraphics[width=\linewidth, trim=0 15 0 10, clip]{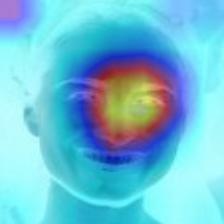}
\end{subfigure}

\rotatebox[origin=l]{90}{~~~~ResNet50}
\begin{subfigure}{.30\linewidth}
    \includegraphics[width=\linewidth, trim=0 15 0 10, clip]{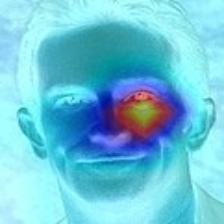}
    \caption{}
\end{subfigure}
\begin{subfigure}{.30\linewidth}
    \includegraphics[width=\linewidth, trim=0 15 0 10, clip]{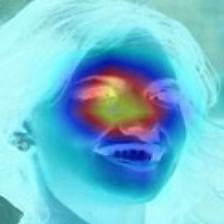}
    \caption{}
\end{subfigure}
\begin{subfigure}{.30\linewidth}
    \includegraphics[width=\linewidth, trim=0 15 0 10, clip]{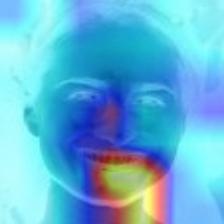}
    \caption{}
\end{subfigure}

\caption{Visual Comparison of Gradient Class Activation Maps~ \cite{jacobgilpytorchcam} between ResNet18 and ResNet50 on images from the CelebA dataset~\cite{liu2015faceattributes}. It is evident that ResNet18 processes more secondary information and utilizes more comprehensible image regions. (HP) refer to head pose estimation model, (G) refer to gaze estimation models.}
\label{fig:resnet18vs50}
\end{figure}

Through a qualitative investigation, as visually depicted in Figure~\ref{fig:resnet18vs50}, it becomes evident that ResNet18 harnesses image information more effectively in different domains. The analysis of Figure~\ref{fig:resnet18vs50} a) reveals that ResNet18 incorporates more extensive facial regions, enabling the integration of nuanced aspects such as head pose. Moreover, pertinent social cues like social smiles and diverse facial expressions find consideration within ResNet18, particularly when ocular recognition poses challenges. It is noteworthy, however, that these considerations are contingent upon the conjunction with ocular data, as evident in Figure~\ref{fig:resnet18vs50} b).
Conversely, ResNet50 at times exhibits an inclination to overly rely on distinct ancillary attributes, inadvertently leading to the underestimation of the significance of ocular information. The ResNet50's focus primarily centers on individual salient indicators for assessing eye contact, yet often without robust integration with other facial features, shown in Figure~\ref{fig:resnet18vs50} c). Notably, these delineated image regions pertaining to the eyes manifest a greater degree of isolation from the broader facial context, which regrettably engenders occasional fallibility in recognition, consequently resulting in the neglect of other pertinent data points.

This observation also indicates that eye contact detection tasks can be accomplished with simple architectures, highlighting the quality of the dataset, which seemingly captures relevant features for eye contact detection in various scenarios and enables efficient training.

\begin{table*}[t]
\begin{center}
\renewcommand{\arraystretch}{1.3}
\rowcolors{2}{gray!10}{gray!30}
\resizebox{\textwidth}{!}{%
\begin{tabular}{l r ccccccc}
  \toprule
 Method & Backbone & \multicolumn{7}{c}{Eye Contact Classification (Accuracy) ↑ [F1-Score] ↑ } \\\cmidrule{3-9}
 &&  OFDIW & DEEPEC & NITEC-WF~\cite{yang2016wider} & NITEC-Gaze360~\cite{gaze360_2019} & NITEC-CelebA~\cite{liu2015faceattributes} & NITEC-Helen~\cite{Le2012InteractiveFF} & NITEC\\
\cmidrule{1-9}
(HP) 6DRepNet~\cite{9897219}-5& ResNet50  &  75.4 [3.6]    & 54.1 [3.1]  & 79.6 [2.4]   &   50.4  [5.9]  &  36.4 [1.3]   &  43.6 [3.9]   & 57.0 [2.7]   \\
(HP) 6DRepNet~\cite{9897219}-15& ResNet50  &  60.9 [28.8]   & 55.1 [33.5]  & 74.5 [22.3]   &   59.5 [48.2]  &  47.9 [42.9]  &  49.9 [35.7]   &60.9 [39.0]  \\

(HP) 6DRepNet~\cite{9897219}-25& ResNet50  & 50.0 [33.8] & 54.7 [49.5]  & 70.8 [37.3] & 61.0 [56.8] & 61.0 [69.4] &  59.0 [61.8]   & 64.9 [59.4]  \\

(G) L2SC-Net~\cite{abdelrahman2022l2csnet}-5& ResNet50                          &  72.0 [18.6]  &  53.4 [24.8]  & 79.1 [15.7]        &   54.5 [21.6] & 46.7 [34.1]      &  51.0 [30.4]    & 61.6 [27.9]         \\ 
(G) L2SC-Net~\cite{abdelrahman2022l2csnet}-15& ResNet50  &  59.4 [39.7]  &59.4 [57.8]   & 74.0 [41.3]   &   68.6 [62.5] & 66.0 [73.9]   &  70.3 [72.2] & 70.1 [64.9]   \\ 
(G) L2SC-Net~\cite{abdelrahman2022l2csnet}-25& ResNet50  &  46.7 [41.2]  &  56.7 [62.5]  & 66.0 [45.1]   & 77.3 [78.9] & 69.5 [79.5] & 73.9 [79.3] & 69.8 [71.1]  \\ 
(G) Gaze360~\cite{gaze360_2019}*-5 &                             RS18-LSTM    & 53.9 [11.0] & 75.1 [3.7]  & 79.2 [4.2] & 49.8 [2.3] & 38.5 [9.4] & 45.9 [12.3] & 57.6 [7.3]\\
(G) Gaze360~\cite{gaze360_2019}*-15 &                               RS18-LSTM      & 56.2 [42.0] & 68.3 [20.6] & 75.3 [25.8] & 57.2 [32.8] & 52.9 [51.9] & 57.0 [50.4] & 63.1 [43.1]\\
(G) Gaze360~\cite{gaze360_2019}*-25 &                              RS18-LSTM       & 57.9 [31.3] & 57.1 [57.2] & 68.6 [35.9] & 63.3 [54.9] & 64.4 [71.8] & 66.9 [71.2] & 66.1 [60.7] \\
Chong~\cite{Chong2020DetectionOE}* &  ResNet50 & 59.3 [47.8] & 68.2 [45.9] & 68.3 [45.9]  & 76.1 [73.9] &  75.3 [79.9] & 81.9 [83.8] &  73.1 [70.2]\\
OFDIW~\cite{Zhang2021OnfocusDI}  &   ResNet18    & 79.3 [33.1] &   67.6 [61.2] & 81.7 [37.2] & 54.4 [19.9] &  74.8 [76.7] &  76.6 [75.4] &  74.3 [61.2]\\

OFDIW~\cite{Zhang2021OnfocusDI} &    ResNet50 & 79.7 [40.0] & 66.5 [59.3] & 80.5 [38.7] & 53.8 [19.2] &  71.7 [73.7] & 74.1 [72.7] &   72.5 [59.0]\\

DEEPEC~\cite{Mitsuzumi2017DEEPEC}&   ResNet18 & 75.3 [10.7] &  67.5 [62.7] &  77.7 [23.9] & 53.4 [27.7] &  51.2 [42.1] &  74.7 [74.5]  &   64.2 [39.9]\\
DEEPEC~\cite{Mitsuzumi2017DEEPEC} &  ResNet50 & 75.3 [10.7] &  70.0 [65.8] & 77.6 [17.7] & 49.8 [22.5] &  50.3 [38.8] & 79.4 [79.5]  & 63.6 [37.1]\\

NITEC (Ours)       &                 ResNet18 & \textbf{80.6} [55.3] &  \textbf{74.3 [73.3]}  &  \textbf{84.3} [59.8] & 87.1 [86.7] & 88.1 [90.3] & 88.6 [89.5] & \textbf{86.4 [83.6]} \\
NITEC (Ours) &                       ResNet50 & 77.8 [53.6] & 72.2 [71.7] & 82.8 [57.0] & 85.1 [84.5] & \textbf{88.3 [90.6]} & \textbf{89.3 [90.5]} & 85.6 [83.0]\\
NITEC (Ours) &                       SWIN-Tiny & 79.0 \textbf{[55.7]} & 74.2 [71.5] & 84.1 \textbf{[60.6]} & \textbf{87.8 [87.8]} & 85.6 [88.1]& 86.7 [87.9] & 85.4 [82.6]\\ 
NITEC (Ours) &                       SWIN-Small& 80.0 [53.9] & 73.0 [70.0]  & 82.9 [57.4] & 81.0 [78.8] & 86.5 [88.7] & 85.9 [87.1] & 84.1 [80.4]\\ 
NITEC (Ours) &                       SWIN-Base & 80.1 [44.1] & 72.4 [67.1] & 83.4 [52.8] & 84.5 [83.1] & 74.3 [75.3] & 80.2 [79.8] & 80.2 [73.0]\\

\bottomrule
\end{tabular}}
\caption{Comparison of different models for eye contact classification, including head pose based and gaze based  estimation methods. The used metrics are accuracy and F1-Score (in square brackets). Models with * are provided by the original authors.}
\label{table:quant_eval_acc}
\end{center}
\end{table*}

In Table~\ref{table:quant_eval_acc}, we conduct another quantitative experiment. Here, instead of average precision, we measure the accuracy along with the F1 Score to include additional non-eye contact models in the comparison. For head pose-based eye contact prediction, we employ the 6DRepnet~\cite{9897219}, a leading model for image-based head pose estimation. For the gaze direction-based approach, we utilize the L2SC-Net~\cite{abdelrahman2022l2csnet}. For further comparison, we include the Gaze360~\cite{gaze360_2019} gaze direction model. Both models were trained on the Gaze360 dataset. Additionally, we include the model proposed by Chong~\etal~\cite{Chong2020DetectionOE} as a comparative benchmark.


For the head pose and gaze direction models, we follow the same procedure as for the Gaze360 data labeling and define predictions as eye contact when the yaw and pitch angles are between -5 and 5 degrees. All other predictions are defined as non-eye contact. In two additional iteration, we increased this threshold to 15 and 25 degrees. 

The results indicate that head pose models are not suitable for eye contact prediction. Although an accuracy of over 70 percent can be achieved on the OFDIW and WIDER FACE subsets, the accuracy should be interpreted with caution, as the label distribution for these two sets are over 70\% (see also table~\ref{table:dataset_comparison}).
The results of the F1-score indicate that neither the recall nor the precision can achieve compatible measures. L2SC achieves similar accuracy results but reaches double-digit values in the F1-score, showing better, yet not satisfying results. The results of Gaze360 resemble those of 6DRepNet at a threshold of 5.
However, when the threshold is increased to 15 or even 25 degrees, all three methods show a significant improvement for the F1-score that saturates between a threshold of 20 and 30 degrees. 
This can be attributed to the fact that the prediction errors of the gaze direction and head pose models are larger than the considered intervals for eye contact. 
This behavior is further analyzed in section~\ref{sec:pred_distri}.

The model by Chong~\etal~\cite{Chong2020DetectionOE } was trained on the largest amount of data by far. This is evident in its consistently robust accuracy and F1-scores. While it outperforms the on OFDIW and DEEPEC trained models on the NITEC test sets, it falls behind both models on the OFDIW and DEEPEC test sets. Especially for the OFDIW datasets, the accuracy and F1-Score remains low, which supports our assumption in section~\ref{sec:cross_eval} about its excessive label noise.

However, for NITEC-trained models based on ResNet and the SWIN-Transformer-tiny~\cite{liu2021Swin} architectures achieve the best results by a significant margin and, thus, can prevail also in this comparison as the most efficient and well-generalized model. We argue that to achieve superior results for the \textit{small} and \textit{base} SWIN architecture, more training data is required than the NITEC dataset currently offers. 


\subsection{In-dataset evaluation}
Table~\ref{table:in_dataset_evaluation} presents the results of our in-dataset evaluation of the NITEC dataset. In this evaluation, we trained the ResNet18 baseline model on the train set of each subset and tested it on all other test sets. For evaluation, we used average precision as the primary metric, supplemented with the F1-Score in parentheses. The model trained on the CelebA subset shows the strongest performance of subsets, as it outperforms not only on its own test set, but also on the WIDER FACE and Helen test set. 
Remarkably, the model trained on the complete dataset surpasses all other models, showcasing the synergistic effect of the composition of data from the different datasets. This highlights the complementary nature of our chosen subsets for NITEC, ultimately resulting in improved generalization performance.

\begin{table}[t]
\begin{center}
\renewcommand{\arraystretch}{1.3}
\rowcolors{2}{gray!10}{gray!30}
\resizebox{\linewidth}{!}{%
\begin{tabular}{l ccccc}
  \toprule
  & \multicolumn{5}{c}{Eye Contact Classification (AP) ↑ [F1-Score] ↑ } \\\cmidrule{2-6}
Train/Test & WF & Gaze360 & CelebA & Helen & NITEC\\
\midrule
WF &\cellcolor{gray!50}52.8 [46.5] & 75.7 [35.5] & 82.1 [56.8] & 87.4 [73.4] & 76.6 [54.0]\\
Gaze360 & 38.8 [35.8] & \cellcolor{gray!50}87.4 [82.7] & 80.5 [66.9] & 84.3 [66.7] & 75.5 [64.6] \\
CelebA & 53.4 [55.0] & 77.4 [52.0] & \cellcolor{gray!50}91.8 [86.0] & 93.0 [86.1] & 81.3 [74.1]\\
Helen & 43.0 [39.8] & 65.5 [36.7] & 78.0 [61.0] & \cellcolor{gray!50}84.1 [74.9] & 69.5 [54.6]\\
NITEC & \textbf{57.0 [59.8]}  & \textbf{93.0 [86.6]} & \textbf{96.0 [90.2]} & \textbf{95.6 [89.5]}  &  \cellcolor{gray!50}\textbf{88.9 [83.6]} \\
\bottomrule
\end{tabular}}
\caption{NITEC subset evaluation based on the ResNet18 model.}
\label{table:in_dataset_evaluation}
\end{center}
\end{table}



\begin{figure*}
\begin{subfigure}{.24\linewidth}
    \includegraphics[width=\linewidth, trim=0 70 0 30, clip]{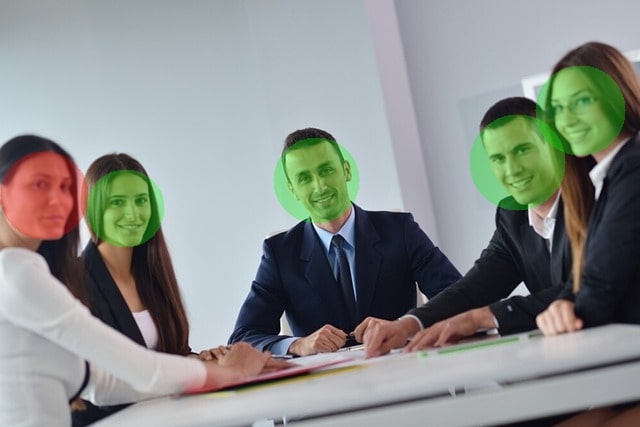}
\end{subfigure}
\begin{subfigure}{.24\linewidth}
    \includegraphics[width=\linewidth, trim=0 70 0 30, clip]{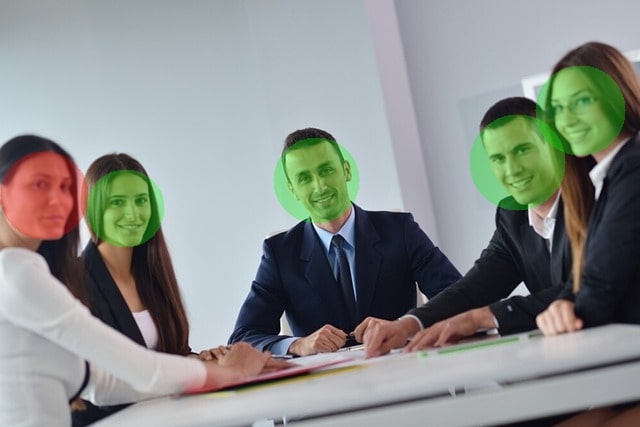}
\end{subfigure}
\begin{subfigure}{.24\linewidth}
  \includegraphics[width=\linewidth, trim=0 70 0 30, clip]{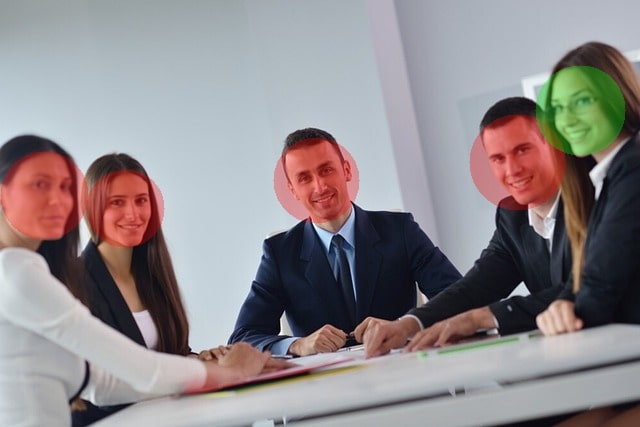}
\end{subfigure}
\begin{subfigure}{.24\linewidth}
   \includegraphics[width=\linewidth, trim=0 70 0 30, clip]{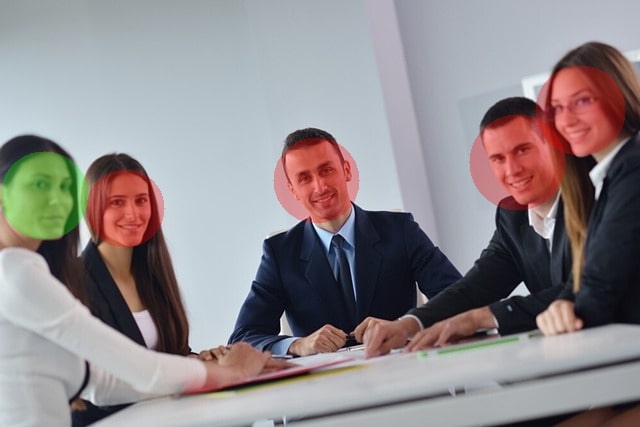}
\end{subfigure}

\begin{subfigure}{.24\linewidth}
    \includegraphics[width=\linewidth, trim=0 380 0 0, clip]{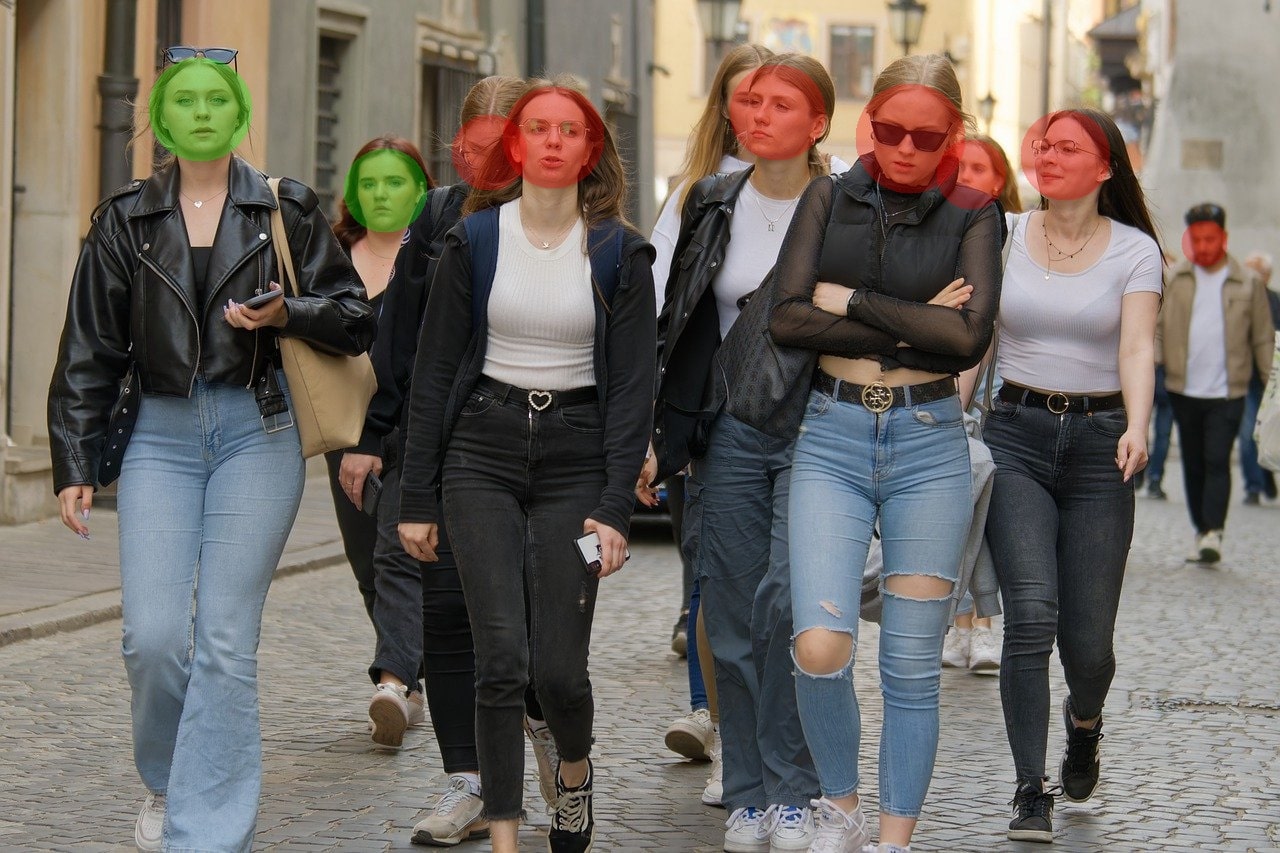}
\end{subfigure}
\begin{subfigure}{.24\linewidth}
    \includegraphics[width=\linewidth, trim=0 380 0 0, clip]{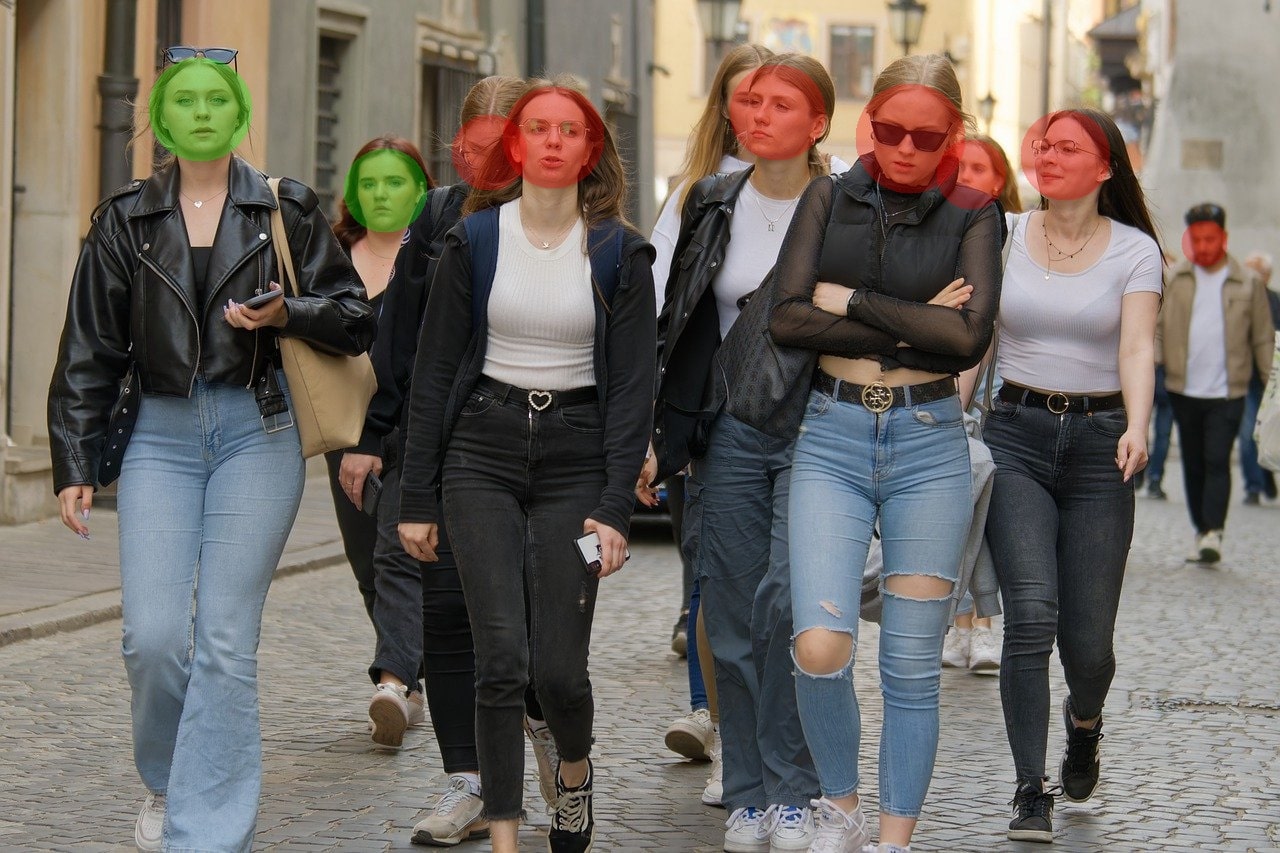}
\end{subfigure}
\begin{subfigure}{.24\linewidth}
  \includegraphics[width=\linewidth, trim=0 380 0 0, clip]{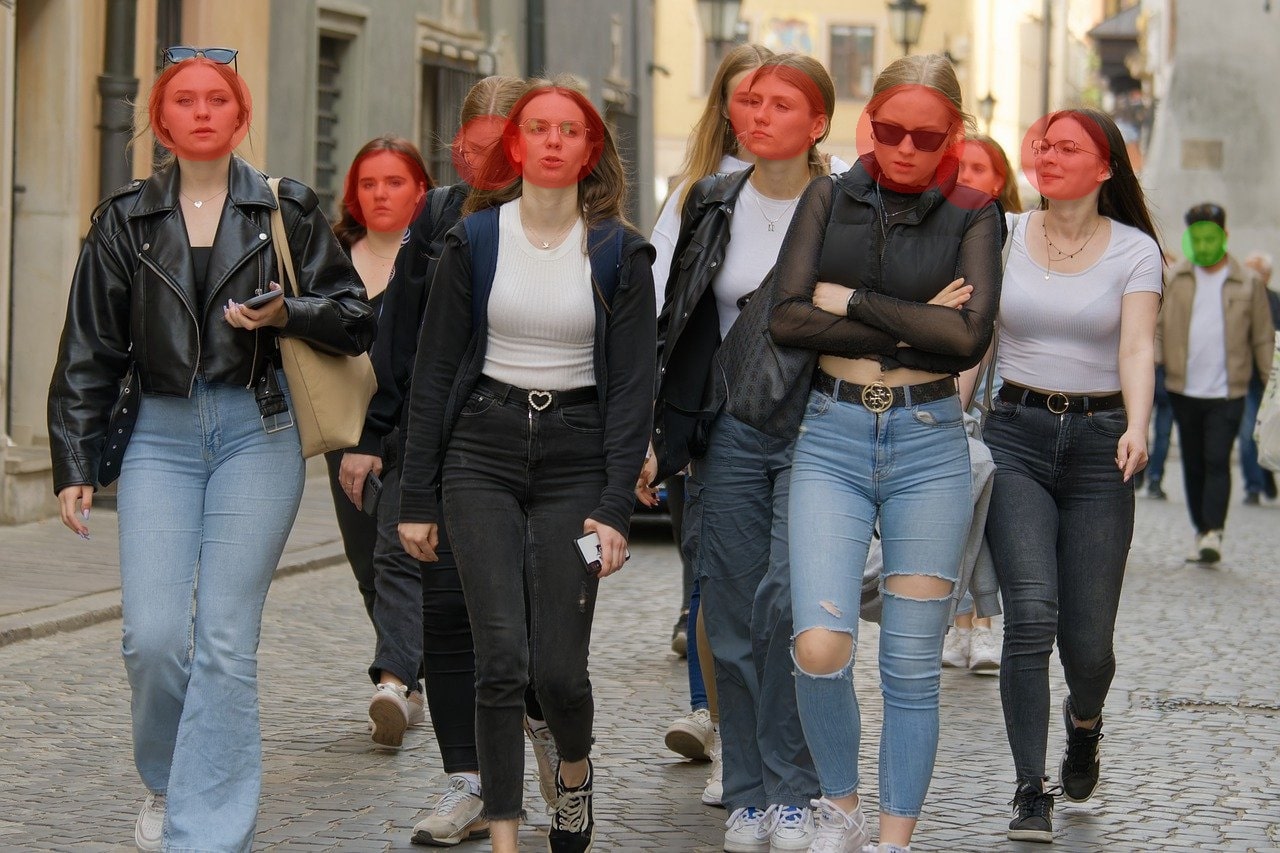}
\end{subfigure}
\begin{subfigure}{.24\linewidth}
   \includegraphics[width=\linewidth, trim=0 380 0 0, clip]{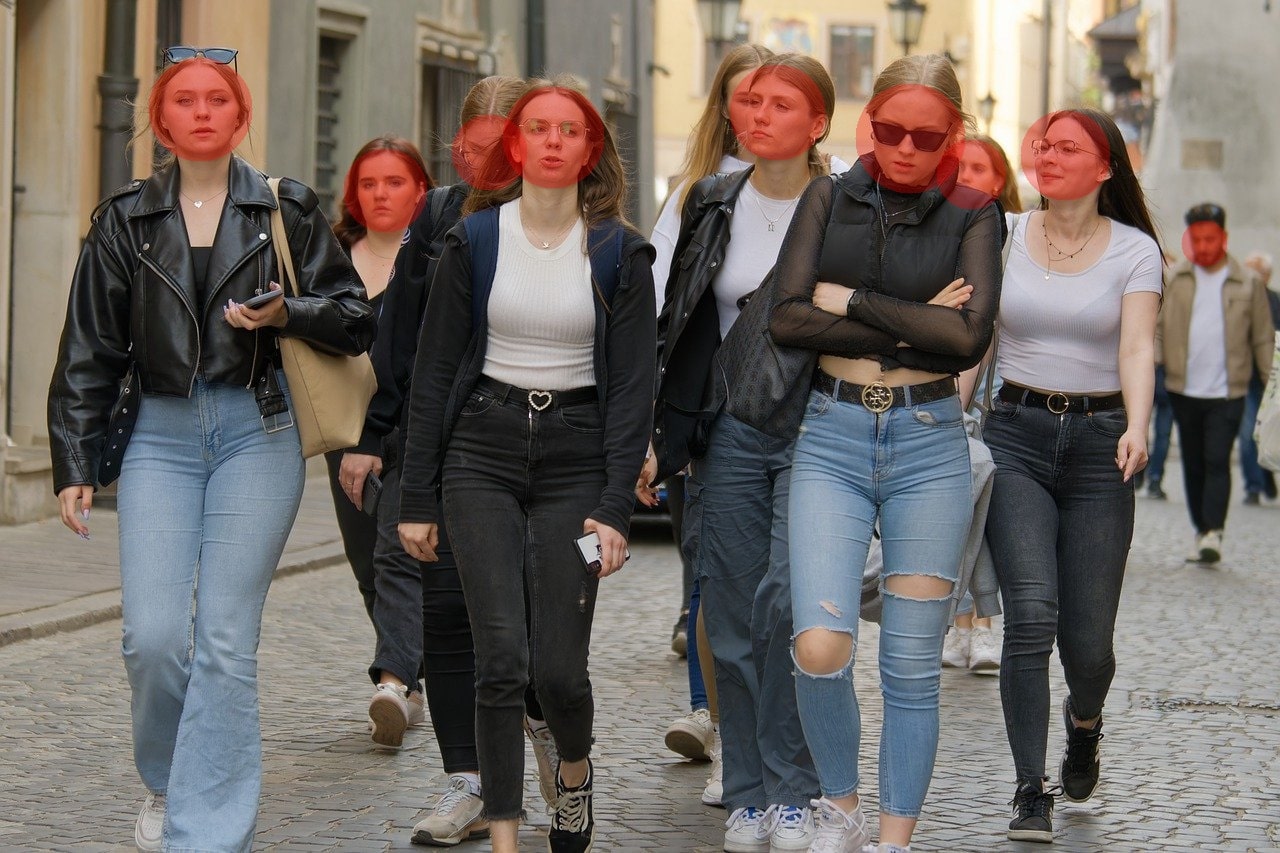}
\end{subfigure}

\begin{subfigure}{.24\linewidth}
    \includegraphics[width=\linewidth]{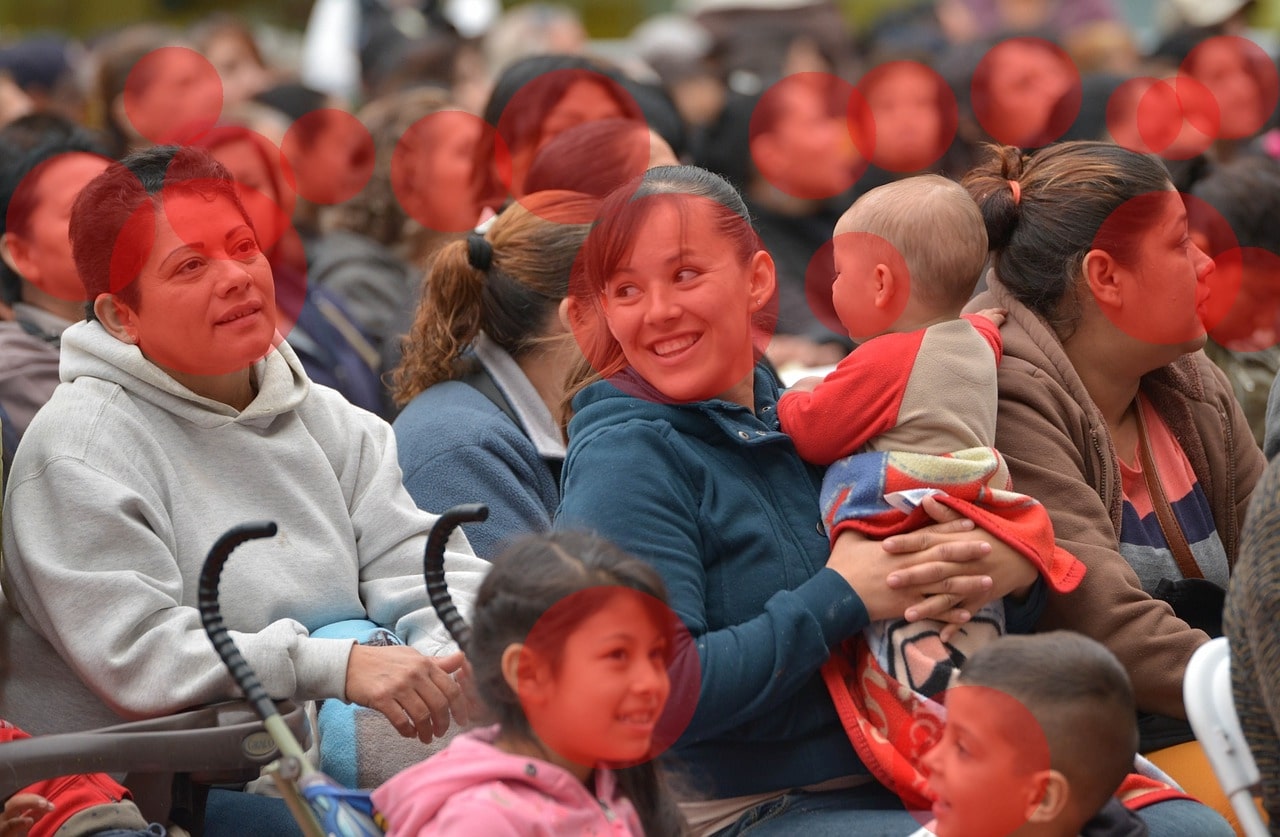}
\end{subfigure}
\begin{subfigure}{.24\linewidth}
    \includegraphics[width=\linewidth]{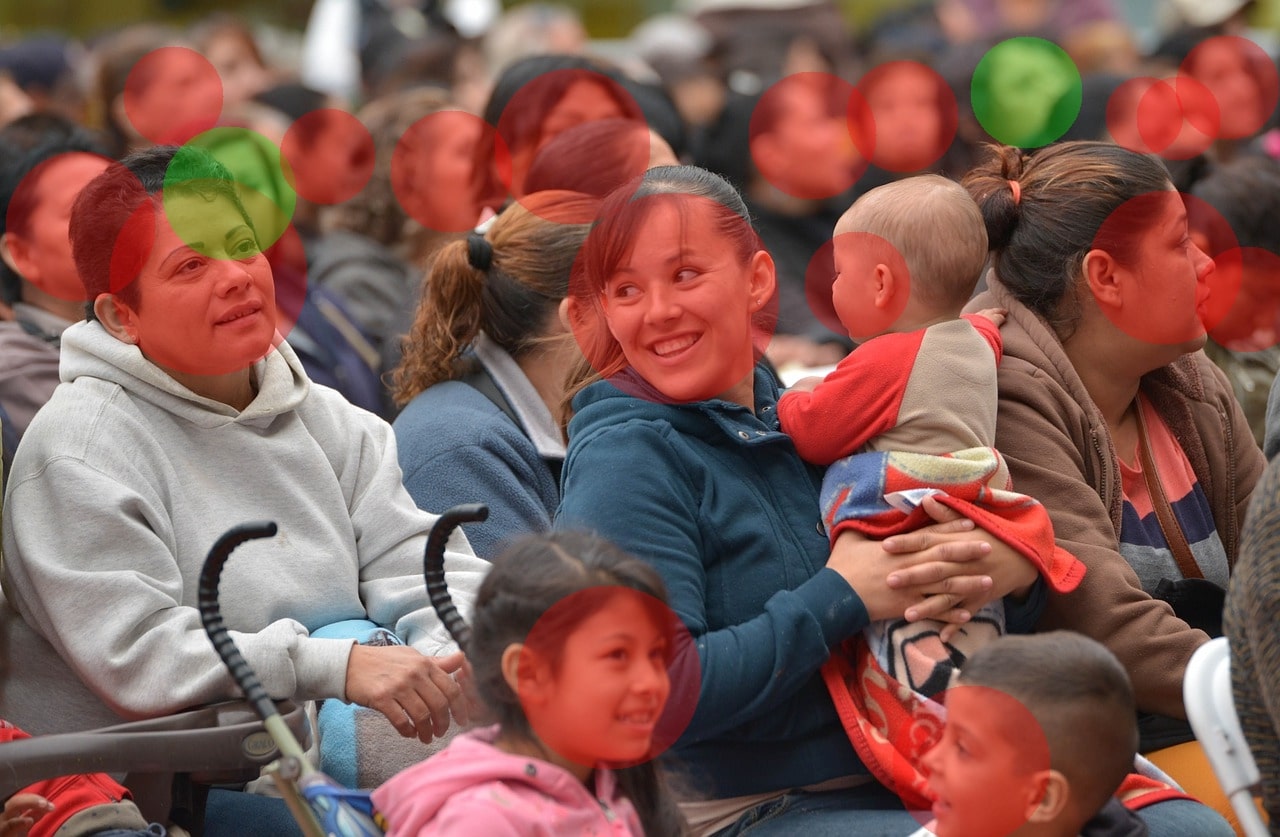}
\end{subfigure}
\begin{subfigure}{.24\linewidth}
  \includegraphics[width=\linewidth]{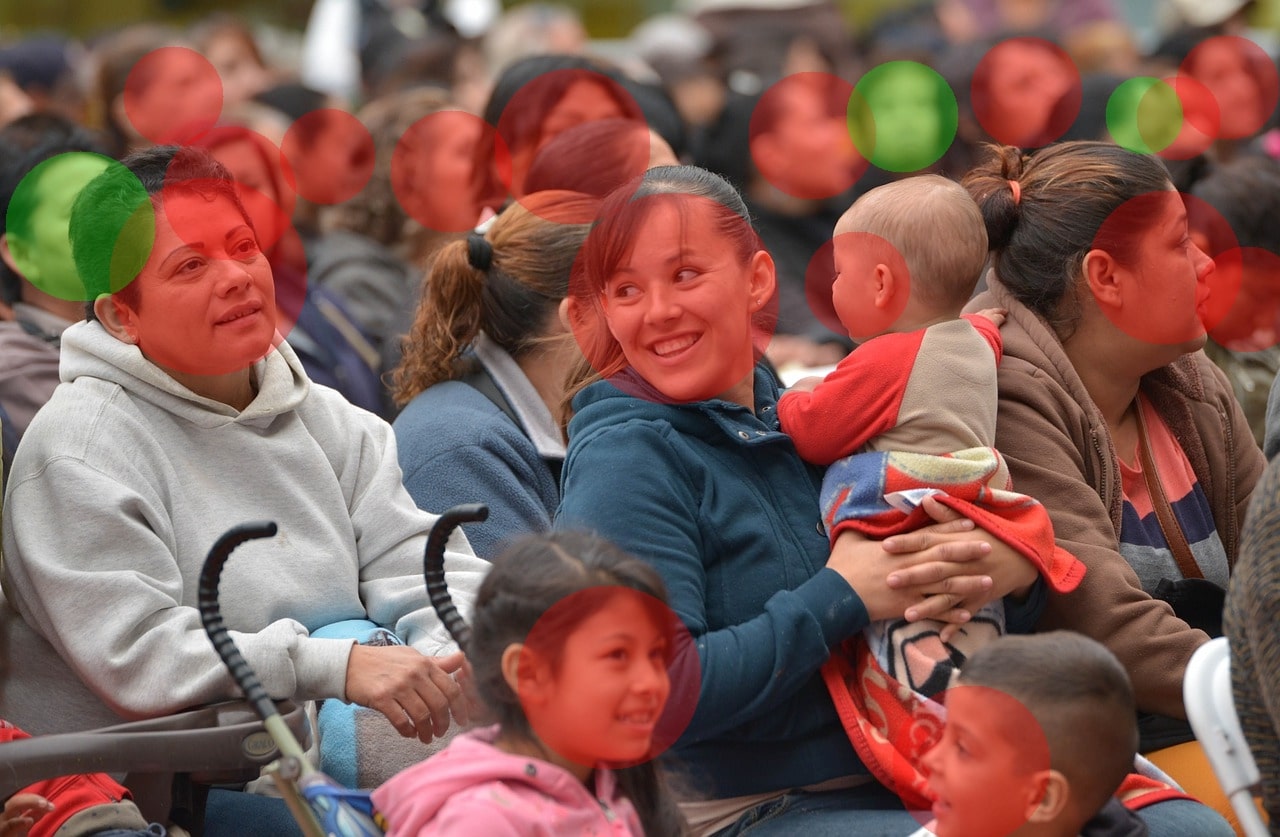}
\end{subfigure}
\begin{subfigure}{.24\linewidth}
   \includegraphics[width=\linewidth]{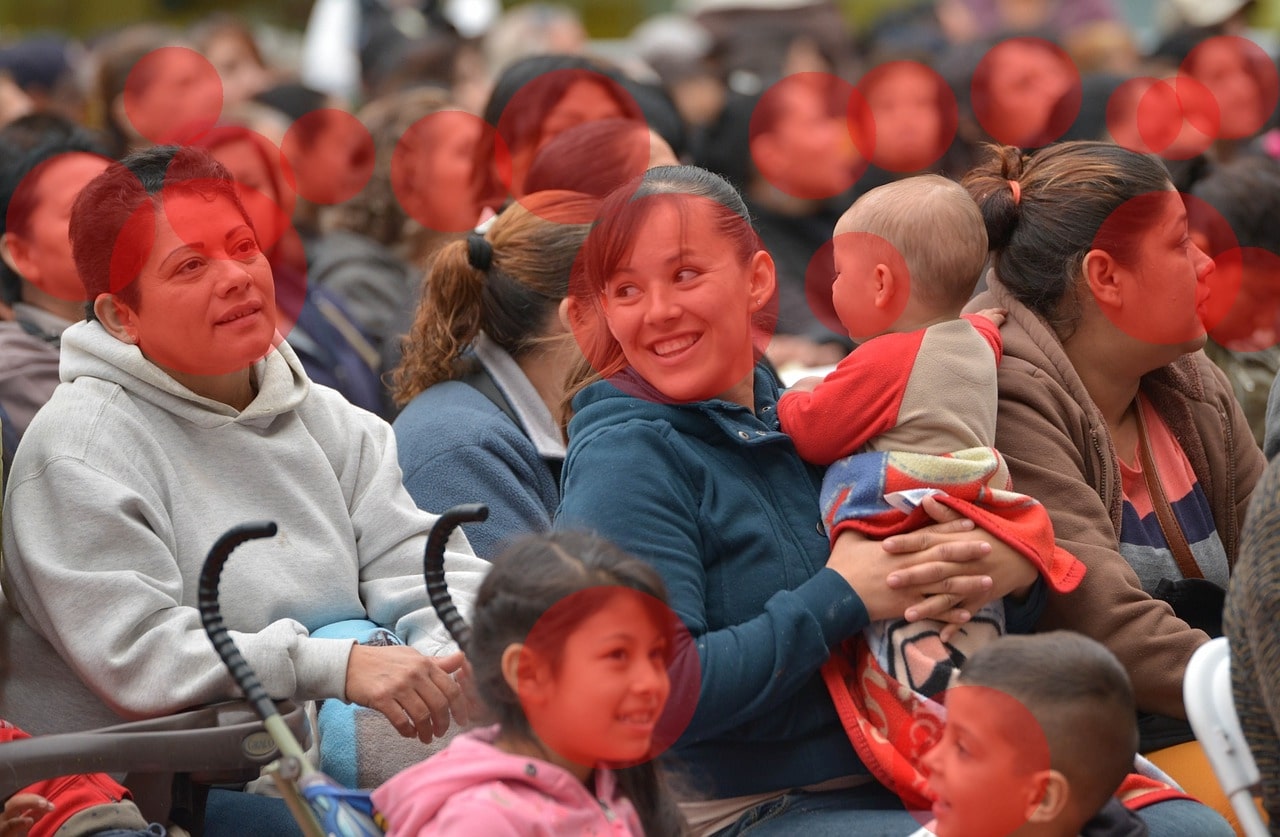}
\end{subfigure}

\begin{subfigure}{.24\linewidth}
    \includegraphics[width=\linewidth, trim=0 440 0 60, clip]{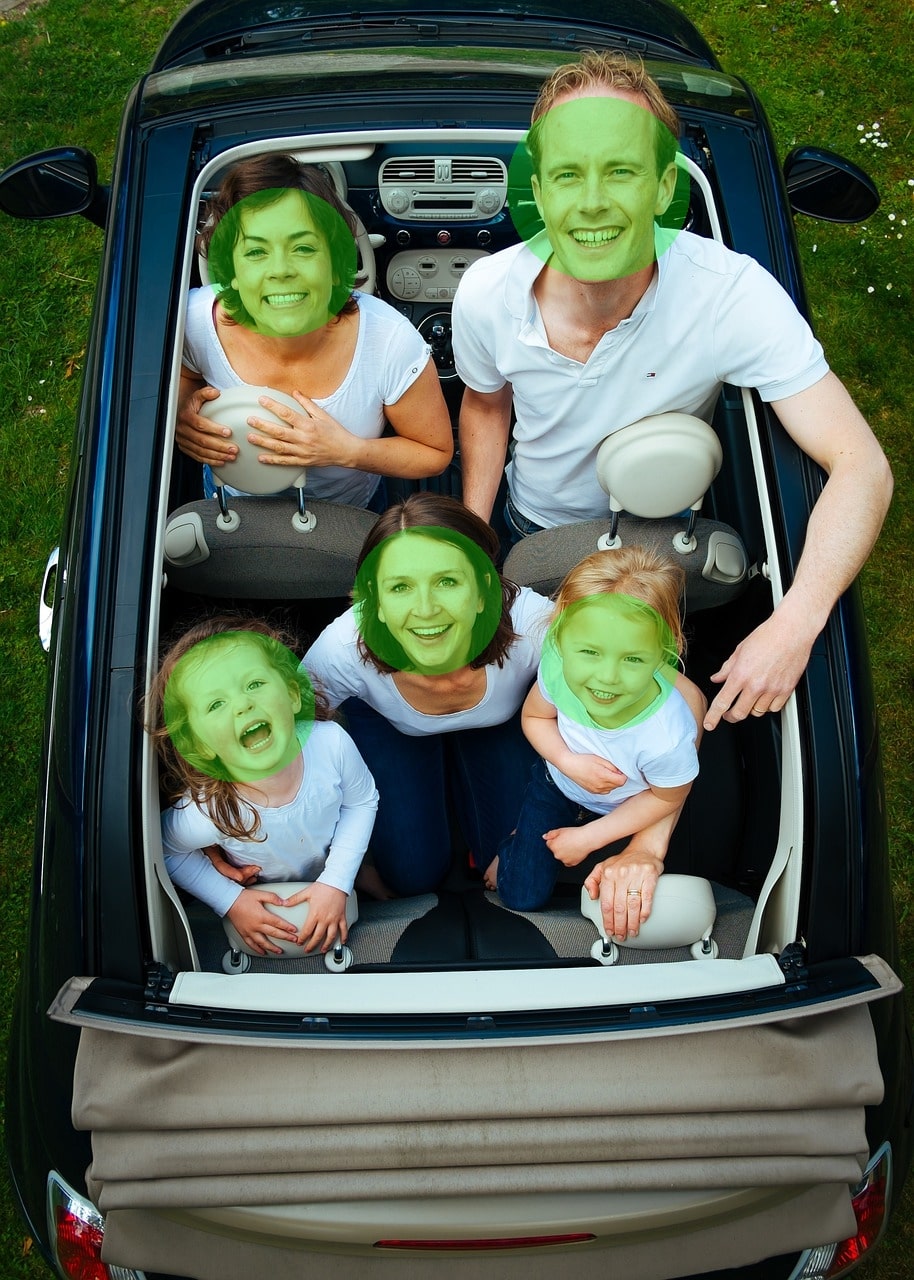}
\end{subfigure}
\begin{subfigure}{.24\linewidth}
    \includegraphics[width=\linewidth, trim=0 440 0 60, clip]{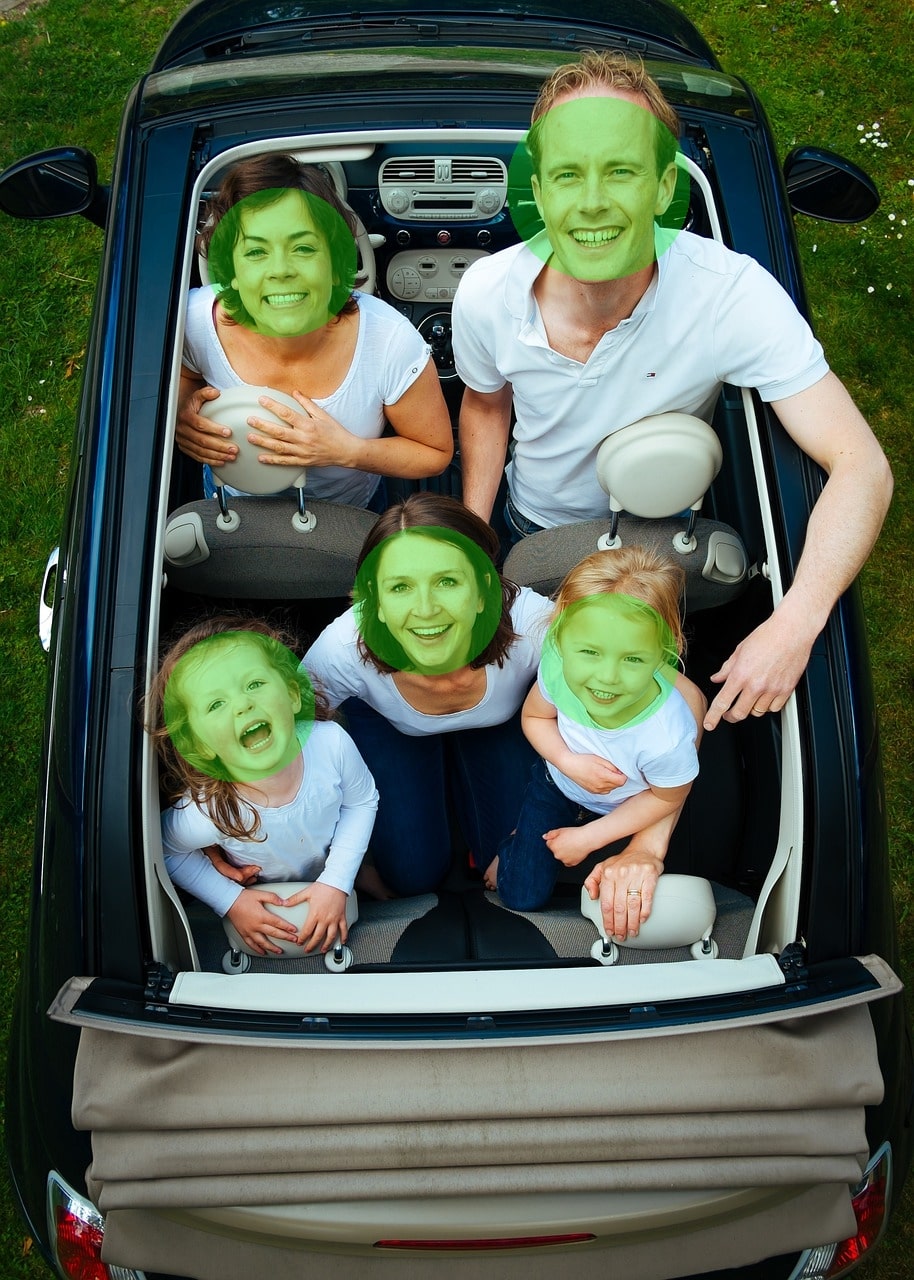}
\end{subfigure}
\begin{subfigure}{.24\linewidth}
  \includegraphics[width=\linewidth, trim=0 440 0 60, clip]{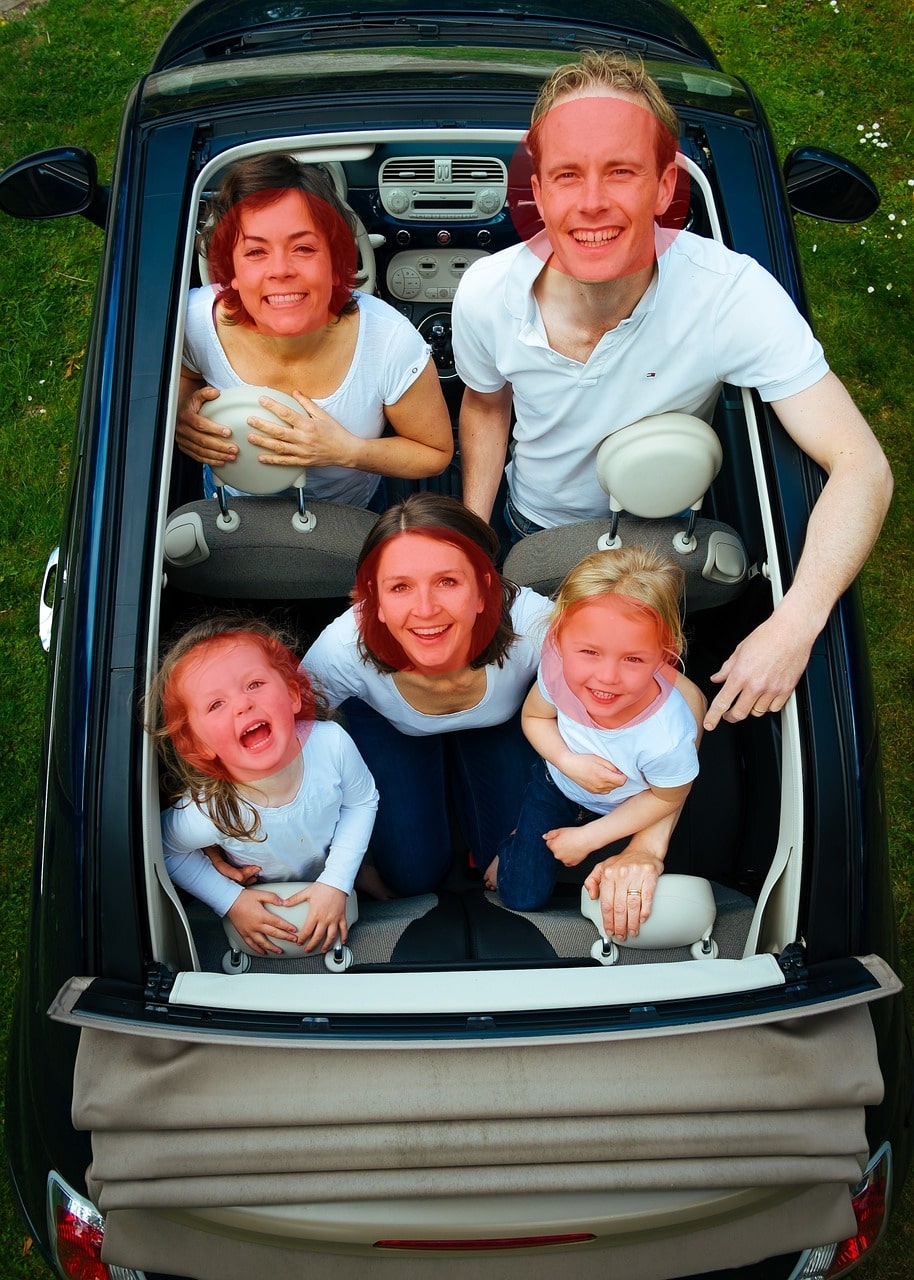}
\end{subfigure}
\begin{subfigure}{.24\linewidth}
   \includegraphics[width=\linewidth, trim=0 440 0 60, clip]{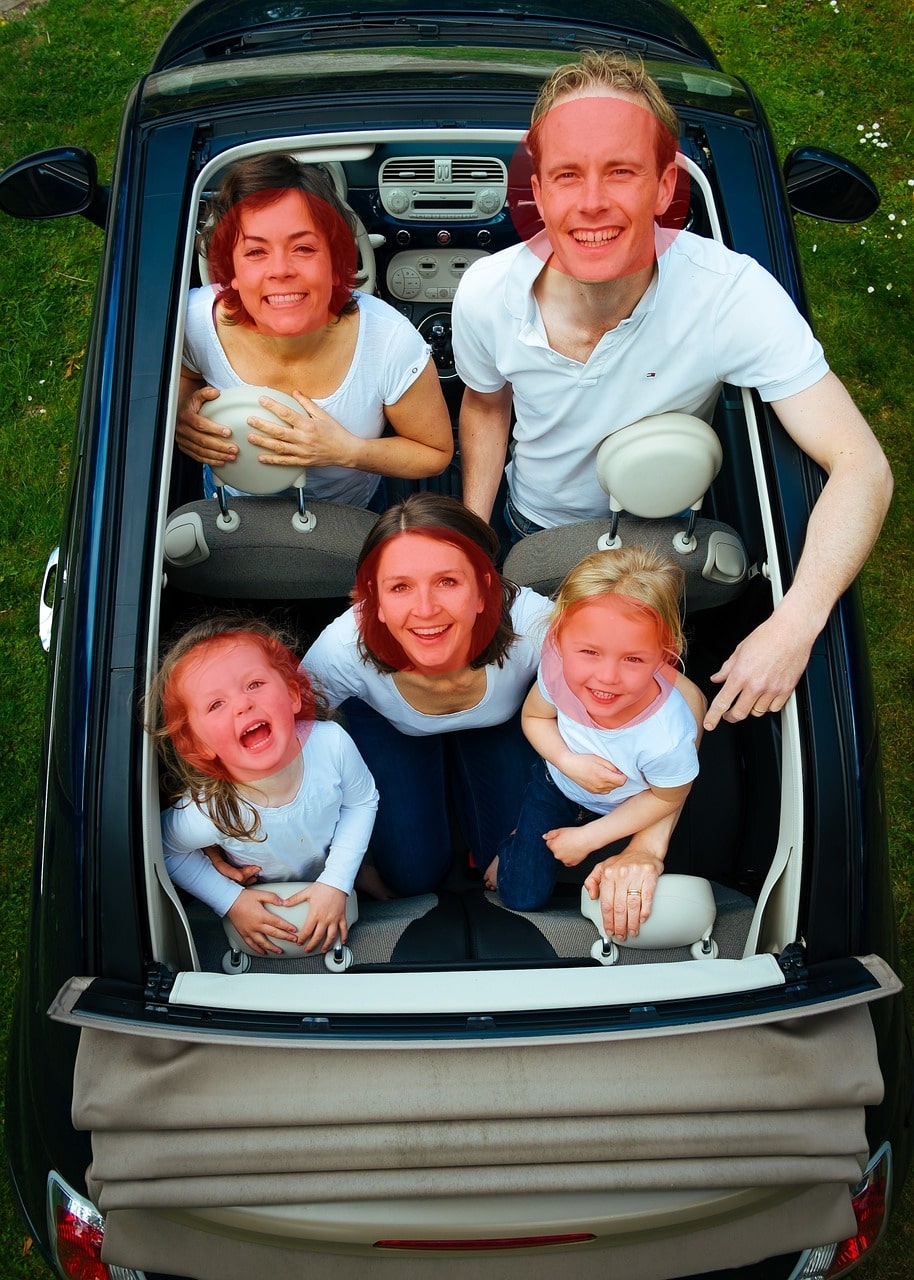}
\end{subfigure}

\begin{subfigure}{.24\linewidth}
    \includegraphics[width=\linewidth, trim=0 140 0 0, clip]{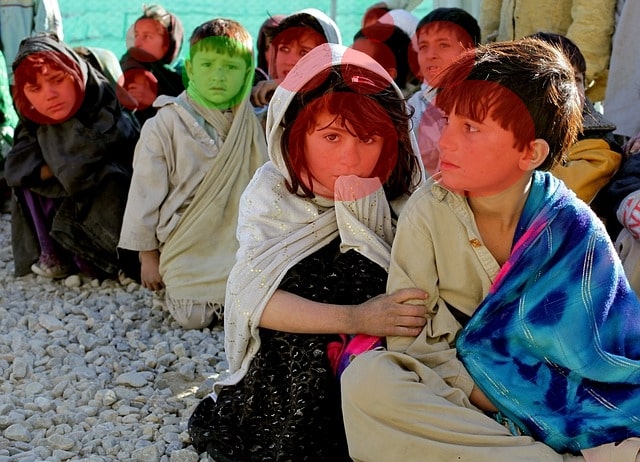}
    \caption{\datasetname}
\end{subfigure}
\begin{subfigure}{.24\linewidth}
    \includegraphics[width=\linewidth, trim=0 140 0 0, clip]{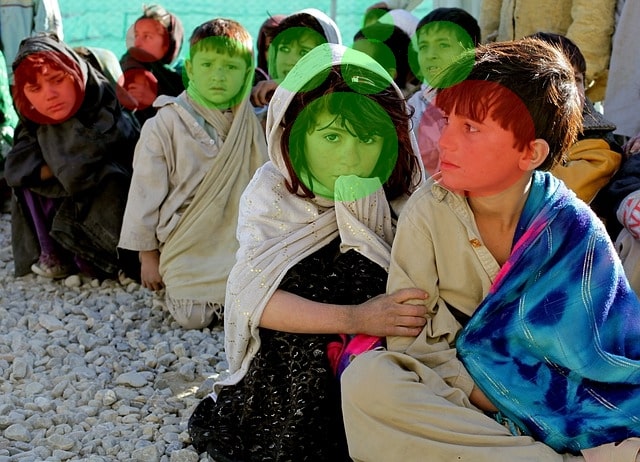}
    \caption{Chong~\etal\cite{Chong2020DetectionOE}}
\end{subfigure}
\begin{subfigure}{.24\linewidth}
  \includegraphics[width=\linewidth, trim=0 140 0 0, clip]{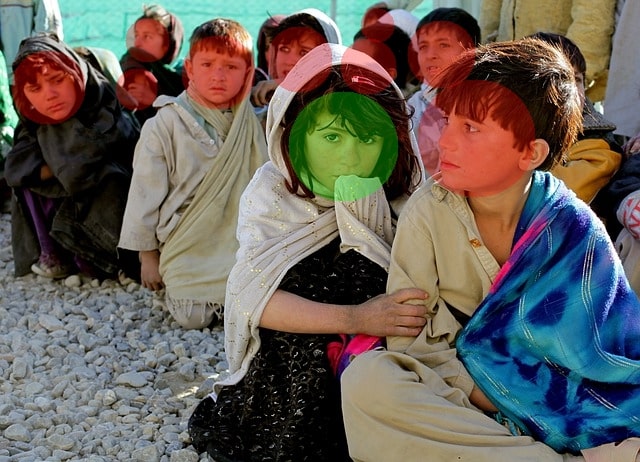}
  \caption{OFDIW~\cite{Zhang2021OnfocusDI}}
\end{subfigure}
\begin{subfigure}{.24\linewidth}
   \includegraphics[width=\linewidth, trim=0 140 0 0, clip]{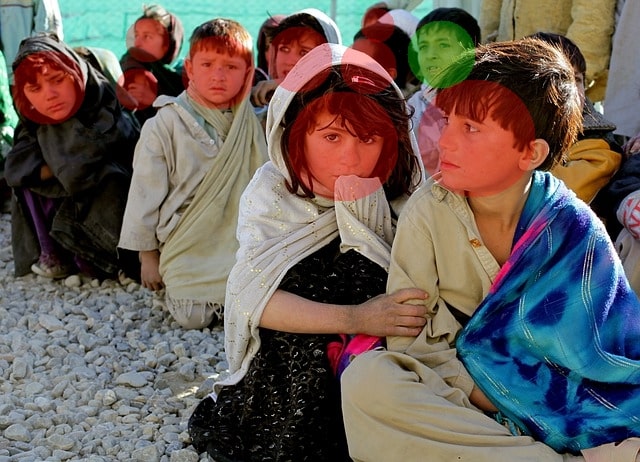}
   \caption{L2SC-Net~\cite{abdelrahman2022l2csnet}}
\end{subfigure}
\caption{Exemplary qualitative results of eye contact classification for NITEC, Chong, OFDIW and the gaze-based L2SC model.}
\label{fig:qual_results}
\end{figure*}
\section{Qualitative analysis}
For qualitative analysis, we five exemplary images and applied the ResNet-18 baseline models for NITEC, OFDIW, as well as the Chong~\etal and the gaze-based L2SC-Net model (with a threshold of 5).
The results are illustrated in Figure~\ref{fig:qual_results}. It exemplifies that OFDIW and L2SC are incapable to detect most of the eye contact faces, while OFDIW even misclassified low-quality faces (second and third row). NITEC and Chong, however, are able to correctly determine the eye contact candidates in row two and four. Particular difference between these two models are shown by more difficult samples given in row three and five. Here, Chong predict False-Positives for heavily blurred faces in the background, while our NITEC model tends for more strict decisions. The reason for this could lie in the choice of WIDER FACE, which we selected with the intention of suppressing potential false positives in challenging images. However, in some cases this can lead to False-Negatives as shown in row five with the girl in the front. This effect is further analyzed in section \ref{sec:pred_distri}.

\begin{figure*}[t]
   \centering
    \includegraphics[width=\linewidth]{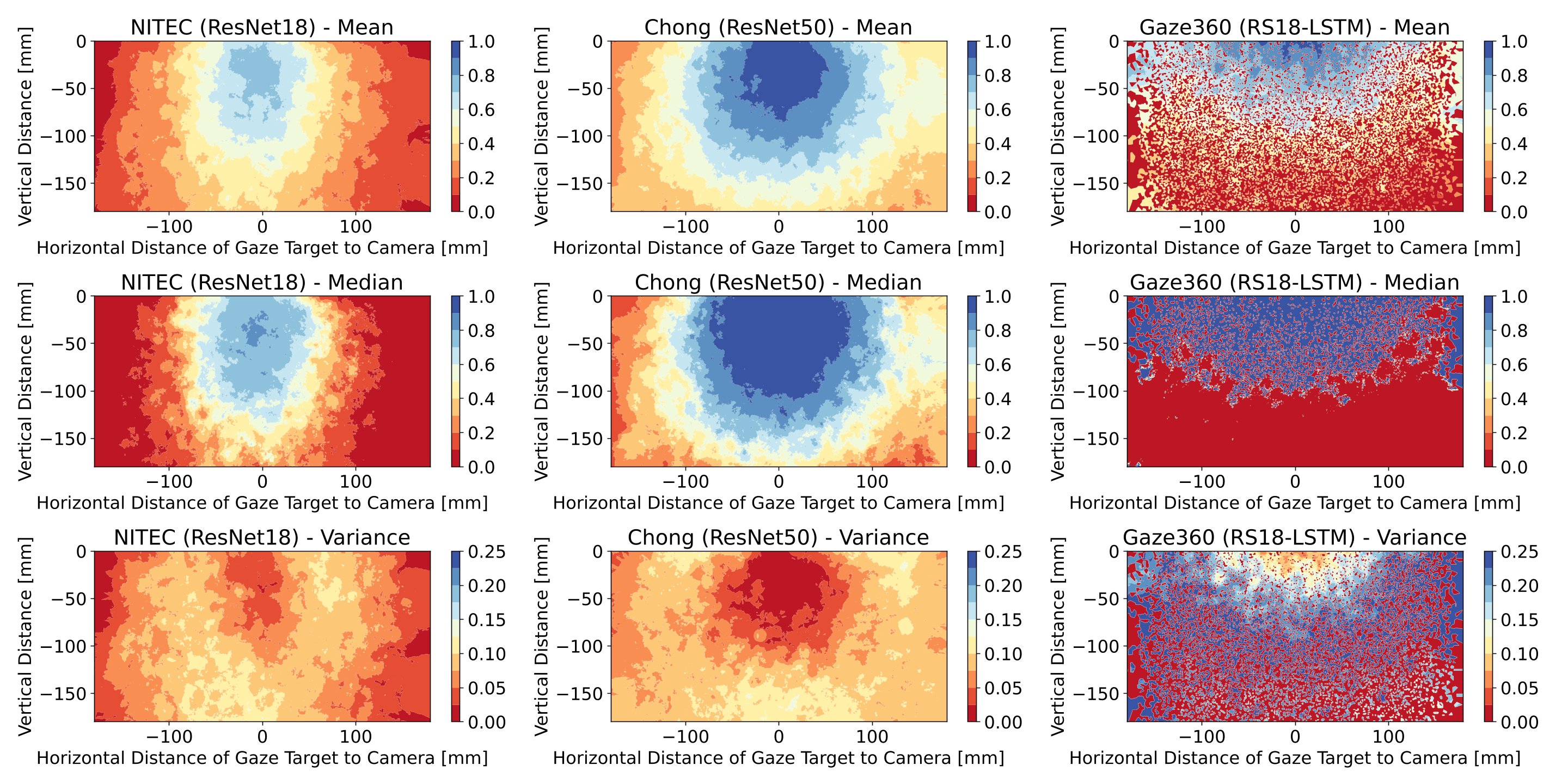}

\caption{Qualitative comparison between different datasets using simple baseline models on the MPIIFaceGaze dataset~\cite{zhang2017s}. The methods considered are Chong~\etal~\cite{Chong2020DetectionOE} with a ResNet50 model, the Gaze360-LSTM trained on the Gaze360 dataset~\cite{gaze360_2019} with a classification threshold range of -15 to 15 degrees to qualify as eye contact, and our NITEC baseline with a ResNet18 model. For the arithmetic mean and the median, a value of 1 indicates predicted eye contact, while 0 indicates no eye contact prediction.}
\label{fig:qualitativAnalysis}
\end{figure*}

\subsection{Prediction distribution analysis}
\label{sec:pred_distri}
Figure \ref{fig:qualitativAnalysis} shows another qualitative comparison of eye contact/non eye contact prediction on the MPIIFaceGaze dataset~\cite{zhang2017s} using baseline models by Chong~\etal~\cite{Chong2020DetectionOE}, Gaze360~\cite{gaze360_2019}, and our NITEC dataset. 
The MPIIFaceGaze dataset consists of 37,788 facial samples, with subjects focusing on the camera level evenly distributed within a relatively small range around the camera, excluding above the camera, resulting in a lack of information in that area. 
Additionally, the subjects have a similar distance from the camera. 
In figure~\ref{fig:qualitativAnalysis}, the predicted values are aggregated using a k-nearest-neighbors algorithm (k=100) and represented with their specific gaze target locations relative to the camera and the prediction values represented in color.
The comparison includes the arithmetic mean, median, and variance. 
When observing the means and medians for the Chong~\etal and our NITEC model, a downward shift of the main region classified as eye contact by the model is noticeable. 
However, examining the Gaze360 model reveals no such shift when considering only gaze direction. This indicates that the discrepancy lies not in the MPIIFaceGaze dataset or its evaluation but rather in the training data of the models. 
This can be explained by perceiving eye contact even when the whole face is observed. 
As the data is hand-annotated, with the eyes located in the upper third of the face, the shift occurs in the region where eye contact is detected.
The graphs for the mean and median also demonstrate that the models gain more confidence as the actual focal point approaches the camera on the horizontal axis and reaches higher values when approaching just below the camera on the vertical axis.
Both the Chong~\etal model and our NITEC model exhibit a uniform decline in predicted eye contact values with increasing distance from the main eye contact region (both in mean and median). 
However, compared to Chong~\etal, our NITEC model is more conservative. 
Only 6.7\% of the values predicted by the NITEC model exceed 0.75, whereas Chong~\etal's model has 35.7\% of the values are above 0.75.
On the other hand, 56.6\% of the values predicted by the NITEC model are below 0.25, whereas Chong~\etal's model has only 7.4\% of the data points predicted below 0.25. 
This allows for greater adaptability of the NITEC model to practical conditions by selecting a threshold for detection.
Comparing the mean and median reveals that the models decisions tend to lean towards the extremes, and the transition between eye contact and non eye contact is slower when considering average values than what the model would predict in the majority of cases, as evident in the subgraphs for the median.
Another important measure for the models generalization capabilities is the scatter in predictions. 
Therefore, the variance within the regions derived from the k-nearest-neighbors algorithm, based on each set of 100 samples, is also shown. 
It can be observed that the variance is very low in the regions where the models assume eye contact (remind NITEC being more conservative than Chong~\etal) and in the regions where no eye contact is predicted. 
As expected, the variance increases in the transitional areas. 
Overall, it becomes apparent that models trained directly on eye contact demonstrate significantly smaller variance, indicating higher robustness than the models trained solely on gaze direction.
Similar to the quantitative analysis, it becomes evident that eye contact detection presents unique challenges that cannot be adequately addressed with existing gaze direction approaches. 
This is primarily attributed to the significant prediction error and the associated variance.
This justifies the need for a dedicated dataset and specialized models for eye contact.
These models offer more flexibility in adjusting the threshold to the specific application field of eye contact and provide significantly higher robustness.

\section{Conclusion}
In this paper, we introduced our hand-annotated NITEC dataset for image-based eye contact detection from egocentric perspective. 
By publicly releasing NITEC we aim to enhance research on nonverbal interaction in the field of human-machine interaction, striving to improve intuitive communication and to reduce misunderstandings. Through multiple quantitative evaluations, we have demonstrated the quality of the dataset, showcasing the exceptional generalization performance even with small baseline models. In future work, we aim to further investigate this behavior and link it with a dedicated user study to gain a better understanding of the subjective perception of eye contact.
\section*{Acknowledgments}
This work is funded and supported by the Federal Ministry of Education and Research of Germany (BMBF) (AutoKoWAT-3DMAt under grant Nr. 13N16336) and German Research Foundation (DFG) under grants Al 638/13-1, Al 638/14-1 and Al 638/15-1.

{\small
\bibliographystyle{ieee_fullname}
\bibliography{bib, egbib}
}

\end{document}